\documentclass[sigconf]{acmart}
\usepackage{natbib}
\usepackage{balance}
\usepackage{amsmath}
\usepackage{booktabs}
\usepackage{tabularx}
\usepackage{adjustbox}
\usepackage{url}
\usepackage{mdwlist}
\usepackage{amsfonts}
\usepackage{multirow}
\usepackage{multicol}
\usepackage{array}
\usepackage{epstopdf}
\usepackage{enumitem}
\usepackage[switch]{lineno}
\usepackage{subfigure}
\usepackage{dcolumn}

\usepackage{graphicx}       
\usepackage{xcolor}         
\usepackage{array}          
\usepackage{booktabs}       
\usepackage{amsfonts}       
\usepackage{caption}        
\usepackage{float}          
\usepackage{bbding} 
\usepackage{enumitem}

\usepackage{algorithm}
\usepackage{algorithmic}
\usepackage{placeins}
\usepackage{amsmath}
\usepackage{adjustbox}
\usepackage{booktabs} 
\usepackage{url}
\usepackage{float} 

\usepackage{stfloats}

\usepackage{newfloat}
\usepackage{listings}
\DeclareCaptionStyle{ruled}{labelfont=normalfont,labelsep=colon,strut=off} 
\floatstyle{ruled}
\newfloat{listing}{tb}{lst}{}
\floatname{listing}{Listing}

\AtBeginDocument{%
  \providecommand\BibTeX{{%
    \normalfont B\kern-0.5em{\scshape i\kern-0.25em b}\kern-0.8em\TeX}}}
    
\copyrightyear{2025}
\acmYear{2025}
\setcopyright{acmlicensed}\acmConference[WWW Companion '25]{Companion Proceedings of the ACM Web Conference 2025}{April 28-May 2, 2025}{Sydney, NSW, Australia}
\acmBooktitle{Companion Proceedings of the ACM Web Conference 2025 (WWW Companion '25), April 28-May 2, 2025, Sydney, NSW, Australia}
\acmDOI{10.1145/3701716.3715259}
\acmISBN{979-8-4007-1331-6/25/04}

\begin{document}

\title{RTBAgent: A LLM-based Agent System for Real-Time Bidding}

\author{Leng Cai}
\orcid{0000-0003-4259-2350}
\authornotemark[0]
\authornote{Both authors contributed equally to this research.}
\email{caileng1923@gmail.com}
\affiliation{%
  \institution{South China University of Technology}
  \city{Guangzhou}
  \country{China}
}

\author{Junxuan He}
\orcid{0009-0000-2365-8794}
\authornotemark[1]
\email{hejunxuan30@gmail.com}
\affiliation{%
  \institution{Shanghai University}
  \city{Shanghai}
  \country{China}
}

\author{Yikai Li}
\orcid{0009-0004-4287-5945}
\email{ykli82086@gmail.com}
\affiliation{%
  \institution{South China University of Technology}
  \city{Guangzhou}
  \country{China}
}

\author{Junjie Liang}
\orcid{0009-0006-7200-0749}
\email{19jjliang22@gmail.com}
\affiliation{%
  \institution{South China University of Technology}
  \city{Guangzhou}
  \country{China}
}

\author{Yuanping Lin}
\orcid{0000-0002-0935-1486}
\email{linyuanping@pazhoulab.cn}
\affiliation{%
  \institution{Pazhou Lab}
  \city{Guangzhou}
  \country{China}
}

\author{Ziming Quan}
\orcid{0009-0006-9816-4290}
\email{dragonquan1112@gmail.com}
\affiliation{%
  \institution{South China University of Technology}
  \city{Guangzhou}
  \country{China}
}

\author{Yawen Zeng}
\orcid{0000-0003-1908-1157}
\email{yawenzeng11@gmail.com}
\authornotemark[0]
\authornote{Corresponding author.}
\affiliation{%
  \institution{ByteDance}
  \city{Beijing}
  \country{China}
}

\author{Jin Xu}
\orcid{0009-0001-8735-3532}
\email{jinxu@scut.edu.cn}
\authornotemark[2]
\affiliation{%
  \institution{South China University of Technology}
  \institution{Pazhou Lab}
  \city{Guangzhou}
  \country{China}
}

\renewcommand{\shortauthors}{Leng Cai, Junxuan He, Yikai Li, Junjie Liang, Yuanping Lin, Ziming Quan, Yawen Zeng \& Jin Xu}

\begin{abstract}
Real-Time Bidding (RTB) enables advertisers to place competitive bids on impression opportunities instantaneously, striving for cost-effectiveness in a highly competitive landscape. Although RTB has widely benefited from the utilization of technologies such as deep learning and reinforcement learning, the reliability of related methods often encounters challenges due to the discrepancies between online and offline environments and the rapid fluctuations of online bidding. To handle these challenges, RTBAgent is proposed as the first RTB agent system based on large language models (LLMs), which synchronizes real competitive advertising bidding environments and obtains bidding prices through an integrated decision-making process. Specifically, obtaining reasoning ability through LLMs, RTBAgent is further tailored to be more professional for RTB via involved auxiliary modules, i.e., click-through rate estimation model, expert strategy knowledge, and daily reflection. In addition, we propose a two-step decision-making process and multi-memory retrieval mechanism, which enables RTBAgent to review historical decisions and transaction records and subsequently make decisions more adaptive to market changes in real-time bidding. Empirical testing with real advertising datasets demonstrates that RTBAgent significantly enhances profitability. The RTBAgent code will be publicly accessible at: \url{https://github.com/CaiLeng/RTBAgent}.

\begin{table}[!h]
\centering
\resizebox{0.47\textwidth}{!}{%
    \renewcommand{\arraystretch}{1.7}%
    \setlength{\tabcolsep}{1.25mm}%
    \begin{tabular}{c|c|c|c|c}
    \toprule
    \textbf{Method}            & \textbf{Scheme}                  & \textbf{Tools}          & \textbf{Adaptability}   & \textbf{Explainability} \\ \hline
    Rule-based Methods & Hyperparameter Tuning  & {\color{red}\XSolidBrush} & {\color{red}\XSolidBrush} & {\color{red}\XSolidBrush} \\  \hline
    RL-based Models          & Model Training          & {\color{red}\XSolidBrush} & {\color{green}\Checkmark}   & {\color{red}\XSolidBrush} \\ \hline
    Chat with LLMs   & LLM API                 & {\color{red}\XSolidBrush} & {\color{red}\XSolidBrush} & {\color{green}\Checkmark}   \\ \hline
    \textbf{RTBAgent (ours)}          & Integrated Intelligence & {\color{green}\Checkmark}   & {\color{green}\Checkmark}   & {\color{green}\Checkmark}  \\ \bottomrule
    \end{tabular}%
}%
\caption{Discussion of three RTB decision methods: Rule-based Methods, RL-based Models, Chat withs LLMs, and our RTBAgent.}
\label{tab:intro}
    \vspace{-1cm}

\end{table}

\begin{figure*}[t]
    \centering
    \includegraphics[width=0.99\linewidth]{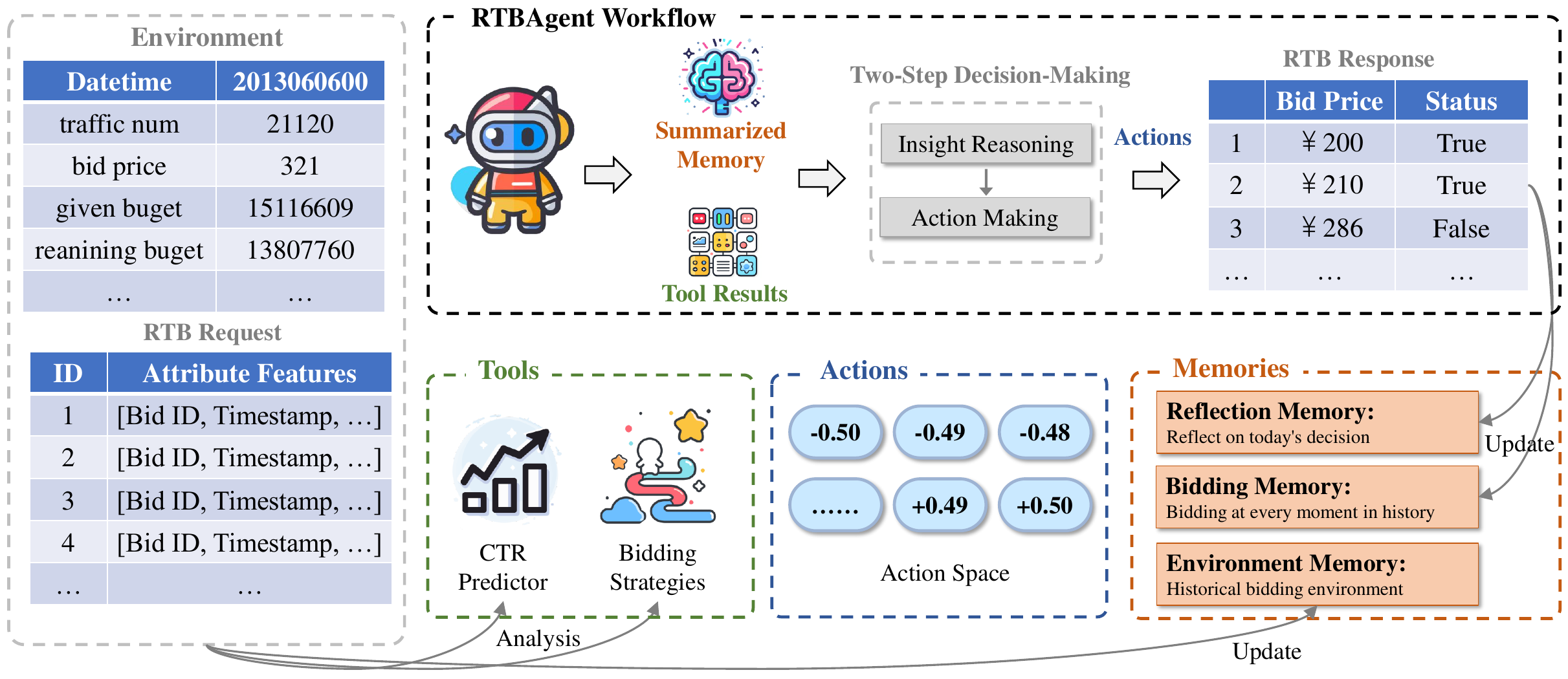}
    \caption{The workflow of our RTBAgent, which is equipped with 4 tools, 3 types of memory, two-step decision-making to execute actions and output bidding prices.}
    \label{fig:framework}
    \vspace{-0.3cm}
\end{figure*}

\end{abstract}

\ccsdesc[500]{Computing methodologies~Artificial intelligence}

\keywords{Real-Time Bidding, Bid Optimization, Large Language Models, Bidding Agents}

\maketitle

\section{Introduction}
The prominence of online advertising within the broader advertising industry is well-established, serving as a pivotal channel for reaching consumers in the digital era. In 2022, the digital advertising industry in the United States achieved a historic milestone, with total revenue exceeding \$200 billion. Programmatic advertising revenue grew by 10.5\% year-on-year, reaching \$109.4 billion, underscoring its expanding significance in the digital landscape~\cite{iab2023internet}. A notable advancement in online display advertising has been Real-Time Bidding (RTB), which enables the real-time buying and selling of ad impressions during a user’s visit. RTB’s main advantage is its ability to automate and scale the purchasing process by aggregating extensive inventory from various publishers~\cite{wang2017display}. It allows for precise targeting of individual users based on real-time behavior, marking a significant shift in digital marketing strategies.

A key challenge within RTB is the development of effective bidding strategies for advertisers. An optimal bidding strategy should promote products to targeted users without disrupting their experience and enhance revenue for publishers. As illustrated in Table \ref{tab:intro}, traditional rule-based bidding strategies are often too rigid and need to adapt to the dynamic nature of the market. While reinforcement learning (RL) approaches~\cite{ding2020challenges} offer better adaptability, they face issues such as the need for extensive training data, difficulties in achieving training convergence, and a lack of interpretability in decision-making, which affects their applicability, trustworthiness, and stability. Consequently, there is a need for more advanced machine-learning models in the RTB domain.

Recent advancements in artificial intelligence, mainly through large language models (LLMs), have introduced innovative solutions to various fields, including knowledge-based question answering~\cite{zhang2023huatuogpt,chen2023soulchat,zhou2024lawgpt,liu2023fingpt,pan2023llms}. However, advertising bidding tasks possess unique competitive characteristics that require heightened awareness and dynamic adjustment. While useful as assistants for bidding and answering, basic LLMs face limitations when directly applied to advertising bidding scenarios. LLM-based agent systems have recently gained attention for their ability to emulate human-like behavior and decision-making. For example, research from Stanford University~\cite{park2023generative} demonstrates how intelligent agent systems can effectively plan and execute complex tasks.  

Our paper introduces \textbf{RTBAgent}, a novel agent framework designed to address the challenges of competitive advertising bidding environments. RTBAgent is equipped with 4 tools, 3 types of memory, and a two-step decision-making process to execute actions. As outlined in Figure \ref{fig:framework}, it simulates real-world advertising agency scenarios and enhances real-time bidding tasks through tools such as CTR predictors and various bidding strategies, integrating expert knowledge with insights into impression value and market conditions. RTBAgent features a versatile multi-memory retrieval system that updates and focuses on relevant data, minimizing noise and adapting swiftly to market dynamics. Its two-step decision-making approach enables it to determine optimal bidding prices in real-time. It has been proven that RTBAgent can increase profitability and has great flexibility in LLM selection. Furthermore, RTBAgent stands out in interpretability, providing transparent insights into its decision-making process, which is a significant advantage over conventional methods that often operate as black boxes. With these capabilities, RTBAgent enables more informed and strategically advantageous decisions, helping advertisers achieve better return on investment in highly competitive markets. The combination of high performance and transparency of RTBAgent can not only be superior to traditional models in effect but also provide a new perspective for the interpretable research of bidding tasks in the field of computational advertising.

Our contributions are summarized as follows:
\begin{itemize}[leftmargin=*]
 \item{To our knowledge, our research proposes the first bidding agency system based on LLMs, aimed at solving the bidding optimization problem in online display advertising under budget constraints.} 
 \item{We innovatively propose a two-step decision-making method for RTB, integrating CTR estimation model, expert strategy knowledge, multi-memory retrieval system, and daily reflection to dynamically adjust bidding strategies to cope with the real-time changing market environment. }  
 \item{Extensive experiments validate that our framework performs exceptionally well across all metrics, achieving a significant overall return.}

\end{itemize}

\section{Related Work}
\subsection{LLM-based Agent Systems}
LLMs are making significant strides towards achieving Artificial General Intelligence (AGI) by enhancing the capabilities of intelligent agents. These models improve autonomy, responsiveness, and social interaction skills, enabling agents to handle complex tasks such as natural language processing, knowledge integration, information retention, logical reasoning, and strategic planning. Recent developments in intelligent agent frameworks, such as AutoGPT~\cite{yang2023auto} and Metagpt~\cite{hong2023metagpt}, have advanced multi-agent collaboration by incorporating standardized operating procedures (SOPs). These frameworks facilitate research by streamlining agent system integration~\cite{huang2023agentcoder,gur2023real,xu2024eduagent,li2024agent,zhang2024finagent}. For instance, EduAgent~\cite{xu2024eduagent} integrates cognitive science principles to guide LLMs, enhancing their ability to model and understand diverse learning behaviors and outcomes. Additionally, Agent Hospital~\cite{li2024agent} utilizes a large-scale language model to simulate hospital environments, enabling medical agents to adapt and improve their treatment strategies through interactive learning.  

\subsection{Bidding Optimization in RTB}
RTB has been a critical focus in online advertising~\cite{wang2015real}, aiming to maximize the value of ad placements within a given budget. Traditional methods employ static parameters~\cite{perlich2012bid,yu2017online,yu2020low} to optimize revenue, often using historical bid data to set bidding parameters. ~\cite{yu2020low} use linear programming to address these optimization problems. Such methods usually fall short in dynamic bidding environments. Researchers have increasingly framed RTB as a sequential decision problem to overcome these limitations, applying RL techniques to enhance automated bidding strategies~\cite{cai2017real,zhao2018deep,wu2018budget,he2021unified}. DRLB~\cite{wu2018budget} approaches budget-constrained bidding as a Markov decision process, offering a model-free RL framework for optimization. USCB~\cite{he2021unified} introduces an RL method that dynamically adjusts parameters for optimal performance, improving convergence rates through recursive optimization. Despite these advancements, RL faces challenges such as training complexity and interpretability, indicating a need for more robust machine learning models in RTB strategies.

\section{Proposed Method}

\subsection{Preliminaries}

\subsubsection{Problem Formulation}

RTB offers various pricing schemes catering to diverse advertiser needs within the online advertising ecosystem. In a second-price auction~\cite{roughgarden2010algorithmic}, the advertiser pays the second-highest bid, denoted as \( c_i \), for the privilege of displaying their ad after winning the bid with \( b_i \) as the highest bidding price. Our study focuses exclusively on the scenario within a second-price auction mechanism. It aims to maximize the achieved objective value under a given budget, as this is the most common business requirement in the industry. W.l.o.g., we consider clicks to be our primary aim value, although other key performance indicators (KPIs), such as conversions, can also be adopted. Thus, the advertiser's strategic challenge is to maximize the cumulative value of clicks, subject to budget constraints. Let \( N \) represent the total number of ad impression opportunities during a specific time period, such as one day, with each impression opportunity indexed by \( i \). The optimization problem can be mathematically expressed as follows:
\begin{small}
\begin{equation}
\begin{aligned}
\max & \sum_{i=1 \cdots N} w_i \cdot v_i \\
\text { s.t. } & \sum_{i=1 \cdots N} w_i \cdot c_i \leq B,
\end{aligned}
\end{equation}
\end{small}where \( v_i \), \( c_i \) represent the value, cost of impression \( i \) respectively. \( w_i \) is a binary value indicating winning or losing the impression \( i \). \( B \) is the total bidding budget. \citet{zhang2016optimal} proves that in a second-price auction, the optimal bid is a function of a scaling factor \( \lambda \), which governs the bid price \( b_i \) as:
\begin{small}
\begin{equation}
\begin{aligned}
b_i = {\lambda} \cdot {v_i}.
\end{aligned}
\end{equation}
\end{small}

Unfortunately, all participating bidders are dynamic, and the auction environment is usually highly non-stationary, which make \(\lambda\) difficult to be determined naively. Thus, the key of our method is to dynamically adjust \(\lambda\) to adapt to the ever-changing market environment.

\subsubsection{Our Solution Paradigm}
RTB can be viewed as a series of random events where each bid is influenced by uncertain factors \cite{amin2012budget}. To address this, we model the RTB scenario using a Markov Decision Process (MDP), a mathematical framework that captures the decision-making process involving probabilistic state transitions. This MDP is characterized by a set of states $\mathcal{S}$ that represent the advertising status of a campaign and an action space $\mathcal{A}$ including the adjustment parameter $a_t$ of the feasible bidding factor $\lambda$. At each time step $t \in \{1, \cdots, T\}$, the agent performs actions $a_t \in \mathcal{A}$ based on the current state $s_t \in \mathcal{S}$ to update $\lambda$ according to its policy $\pi: \mathcal{S} \mapsto \mathcal{A}$. The state then transitions to a new state according to the transition dynamics $\mathcal{T}: \mathcal{S} \times \mathcal{A} \mapsto \Omega(\mathcal{S})$, where $\times$ represents cartesian product and $\Omega(\mathcal{S})$ is a set of probability and distributions over $\mathcal{S}$. The environment provides an immediate reward to the agent based on a function of the current state and the agent's actions, denoted as $r_t: \mathcal{S} \times \mathcal{A} \mapsto \mathcal{R} \subseteq \mathbb{R}$, where $\mathbb{R}$ is a reward space.  

To this end, the objective is to discover a policy \(\pi\) that links states to actions, intending to maximize the total discounted reward within a set time frame, all while considering budget limitations. The policy \(\pi^*\) that we seek is the one that maximizes the expected cumulative reward, as shown in Eq.(3):

\begin{small}
\begin{equation}
\begin{aligned}
\pi_\theta^* = \arg\max_{\pi_\theta} \mathbb{E_{\pi_\theta}}\left[ \sum_{i=0}^{T} \gamma^{i} r_{t+i} \mid s_t=s \right].
\end{aligned}
\end{equation}  
\end{small}We extend this optimization challenge to our RTBAgent, where the policy \(\pi_\theta^*\) is defined as:

\begin{small}
\begin{equation}
\begin{aligned}
\pi_\theta^* = \arg \max_{\pi_\theta} \mathbb{E}_{\pi_\theta}\left[\sum_{i=0}^T \gamma^i r_{t+i} \mid s_t = s, \rho_t = \rho\right],
\end{aligned}
\end{equation}  
\end{small}where $\rho(\cdot)$ is a specialized module that encapsulates beneficial internal reasoning processes and the action policy for the RTBAgent is given by:

\begin{small}
\begin{equation}
\begin{aligned}
&\pi_{\text{RTBAgent}}\left(a_t \mid s_t,\rho_t \right) = G\left(s_t, \rho_t \right) \\
&\rho_t = \rho\left(s_t, F_{{t}}^{sum}, F_{{t}}^{tool}, F_{{t}}^{ins}, F_{{t}}^{act}, F_{{t}}^{ref}\right), 
\end{aligned}
\end{equation}  
\end{small}where \(G(\cdot)\) is a operation parsing function used to perform compatible formal operations in the environment. The RTBAgent, powered by LLMs, refines the inference information \(\rho_t\) to include various operations, i.e., the summary of memories \(F_{{t}}^{sum}\), tools \(F_{{t}}^{tool}\), insights \(F_{{t}}^{ins}\), actions \(F_{{t}}^{act}\) and reflections \(F_{{t}}^{ref}\), which will be explained specifically later.

Due to the inherent limitations of LLMs, such as the lack of dominance in continuous value output and insensitivity to numbers, our research refines the operation based on the basic factor $\lambda_{base}$ obtained from expert bidding strategies. The optimal scaling factor $\lambda_{t}$ at each time step $t$ is determined through the adjustment action $a_{t}$ provided by the strategy \(\pi\), i.e., $\lambda_{t} = \lambda_{base} \cdot (1 + a_{t})$. Integrating these operations allows the RTBAgent to continuously interact with the bidding environment during training, driving it towards the optimization goal as expressed in the following equation:  

\begin{small}
\begin{equation}
\begin{aligned}
 &\pi_{\text{RTBAgent}}^* = \arg \max_{\pi} \mathbb{E}_\pi\left[\sum_{i=0}^T \gamma^i r_{t+i} \mid s_t = s, \rho_t = \rho\right] \\
 & \text{s.t. } \pi(a_t\mid s_t, \rho_t) = G\left(s_t, \rho_t \right)  \enspace with \enspace Eq.(5) \enspace \forall t,
\end{aligned}
\end{equation}
\end{small}where the reward of RTBAgent comes from the comprehensive information of decision results and self-reflection, achieving self alignment\cite{yuan2024self}. This approach ensures that RTBAgent's actions are meticulously aligned with the policy that delivers the highest expected return on bids, taking into account the current state and decision insights at each juncture.    

\subsection{Overall Framework}
The RTBAgent framework, as shown in Figure \ref{fig:framework}, mirrors the operational structure of a real-world bidding firm. It integrates a comprehensive set of bidding analysis tools $\mathcal{H}$, alongside a well-defined profile, action set $\mathcal{A}$, and memory set $\mathcal{M}$. The RTBAgent is guided by a configuration file that imbues it with the acumen of a bidding specialist, grounded in a context crafted for its role. It leverages bidding analysis tools to meticulously evaluate the potential value of ad impression requests, drawing on current bidding conditions and request data to proffer expert-guided bidding strategies. The memory module is designed to offer a robust, multi-dimensional retrieval system. It segments and updates information incrementally, ensuring the agent has access to accurate and relevant data. Additionally, the RTBAgent features a reflection module that it uses for regular decision reviews, thereby cultivating valuable insights for enhancing subsequent actions. Central to the RTBAgent is the action module, which employs a two-step decision-making process to formulate and execute well-considered actions. This process is pivotal for determining the optimal bidding price, ensuring that the agent's actions are strategically aligned with the goal of achieving the highest return on investment in the bidding process. 

\subsection{Environment}
The dynamics of a bidding environment are characterized by the continuous emergence of new data following each round of bidding, which is a concrete representation of the current state $s_t$. This data encompasses a variety of metrics, such as the current volume of bids, historical success rates in securing bids, prevailing market prices, the average cost per bid, the total budget allocated for bidding, and the remaining budget. These elements are crucial as they provide a comprehensive snapshot of the bidding landscape at any given time. Including such detailed environmental information is instrumental for the RTBAgent's decision-making process. It allows the system to not only assess the immediate bidding scenario but also to anticipate and adapt to potential shifts in the market dynamics. This real-time analysis and understanding of the bidding environment are essential for the RTBAgent to make informed and strategic decisions.

\subsection{Components of Agent}
\label{sec:agent}

\subsubsection{Profile} 
To help LLMs understand the bidding process, we define the profile of our RTBAgent as follows,
\vspace{-0.2cm}
\begin{center}
\fcolorbox{black}{gray!10}{\parbox{0.97\linewidth}{
You are a senior data analyst specializing in in-depth research and strategy development in the field of real-time bidding (RTB) advertising placement. You use advanced data analysis tools and algorithms to guide advertisers to gain an advantage in fierce market competition...
}}
\end{center}
by presenting the problem background, its own role, and action goals in the form of text, it can better perform the reasoning process for specific tasks.
\subsubsection{Tools}

We incorporate a various tool set $\mathcal{H}$ for RTBAgent, including a click-through rate (CTR) prediction model and bidding decision strategies. We use Factorization Machines (FM)\cite{rendle2010factorization} as the CTR prediction model, which is widely used and can estimate the value for each impression stream. A separate CTR prediction model is trained for each advertiser within this framework. Additionally, we use a series of rule-based strategies to complement expert knowledge, including MCPC~\cite{lee2018estimating}, LIN~\cite{perlich2012bid}, and LP~\cite{dantzig2002linear}, which will be further discussed in the \textbf{\nameref{sec:overallcomparison}} section. These methods are widely used in the industry, based on prior knowledge, and demonstrate high referential value in an offline environment. It is important to note that in the bidding decision-making process, we only need to base our decision on one of the bidding decision models to assist in the decision-making. 

\subsubsection{Actions} 
Due to the uncontrollable and unreliable nature of generative LLMs in predicting consecutive bidding prices, we suggest changing this process to allow the LLMs to predict an adjustment factor. Specifically, we define a adjustment space with an observation range from -0.5 to 0.5. This enables the RTBAgent to adjust the expert knowledge suggested decisions based on the current state and historical decisions in order to better adapt to the dynamic environment.

\subsubsection{Memories} 
In the RTBAgent, we design three types of memory: environment memory $\mathcal{M}^{env}$, bidding memory $\mathcal{M}^{bid}$, and reflection memory $\mathcal{M}^{ref}$, where $\{\mathcal{M}^{env}, \mathcal{M}^{bid}, \mathcal{M}^{ref}\} \subset   \mathcal{M}$. Specifically, $\mathcal{M}^{env}$ stores the market environment after each decision, allowing the RTBAgent to refer to historical data to make wiser decisions when facing new bidding opportunities. $\mathcal{M}^{bid}$ records the bidding behaviors and reasons in different market environments. By analyzing $\mathcal{M}^{bid}$, the RTBAgent can identify which strategies are more effective in specific situations, optimizing and adjusting future bids. Additionally, $\mathcal{M}^{bid}$ can help the RTBAgent recognize potential patterns and trends, such as specific bidding strategies that are more likely to succeed under certain conditions. $\mathcal{M}^{ref}$ is the RTBAgent's self-assessment mechanism, recording the reflection process and results after each decision. $\mathcal{M}^{ref}$ enables the RTBAgent can to understand why some decisions did not achieve the expected effects, thus avoiding similar mistakes in the future. The core of $\mathcal{M}^{ref}$ lies in continuous learning and improvement, ensuring that the RTBAgent can maintain competitiveness in the ever-changing market environment.

\subsection{Workflow of RTBAgent}
This section elucidates the operational sequence of the RTBAgent, encompassing three pivotal stages: information gathering, two-step decision-making, and daily reflection.

\subsubsection{Information Gathering}
At the heart of RTBAgent's functionality is the aggregation of pertinent data. It should be noted that each action and environmental feedback record will be saved in real time to generate memory. Throughout the bidding process, extensive logs are generated, encompassing decisions, reflections, and environmental contexts. The synthesis of these logs into a summarized memory is pivotal for informed decision-making. The summarized memory at any given time step \( t \), denoted as \( F_{{t}}^{sum} \), is formulated by integrating information from bidding, reflective, and environmental memories:
\begin{small}
\begin{equation}
\begin{aligned}
F_{{t}}^{sum} = \sum_{{i=bid,env,ref}} LLM(\varphi^{{sum}}(m^i_{t})),
\end{aligned}
\end{equation}
\end{small}where $\sum{}$ is a concatenation operation of multiple strings, $\varphi^{sum}(\cdot)$ is a prompt template for information gathering and $m^i_{t} \in \mathcal{M}^i$ is the memory for the type $i$ at time step $t$.

Additionally, the RTBAgent leverages a suite of tools, encapsulated within \( F_{{t}}^{tool} \), to provide foundational insights for two-step decision-making. Central to this are the CTR estimations, represented by \( V_t \), and strategic bidding recommendations, denoted as \( {\lambda}_{base} \). Therefore, the output of the tool can be represented by a pair (\( V_t \), \( {\lambda}_{base} \)). Here, \( V_t = \{v_t^1, v_t^2, \cdots, v_t^{d_t}\} \) is derived from a trained predictive model, forecasting the potential user engagement by the impression feature vector \( X_t = \{x_t^1, x_t^2, \cdots, x_t^{d_t}\} \), while \( {\lambda}_{base} \) is algorithmically determined based on historical data, and \( d_t \) is the number of impressions at the current time step \( t \), and satisfies the following:
\begin{small}
\begin{equation}
\begin{aligned}
\sum_{t=1}^{T} d_{t}=N,
\end{aligned}
\end{equation}
\end{small}where $N$ is the total number of ad impression opportunities during one day.
\subsubsection{Two-Step Decision-Making}
The RTBAgent's two-step decision-making process plays a crucial role in its strategic capabilities. The first step involves insight reasoning, represented by \( F_{t}^{ins} \), and is responsible for analyzing potential decision ranges, including their benefits, drawbacks, possible outcomes, and associated risks. This step can be formulated as follows:
\begin{small}
\begin{equation}
\begin{aligned}
F_{t}^{ins} = LLM(\varphi^{{ins}}(s_t, F_{t}^{sum}, {\lambda}_{base})),
\end{aligned}
\end{equation}
\end{small}where $\varphi^{ins}(\cdot)$ is a prompt template for insight reasoning.

Following the insight reasoning, the action making step, represented by \( F_{{t}}^{act} \), is where the actual bidding action \( a_t \) and its reason \( {rea}_{t} \) are determined. The output of this step is a binary tuple \( (a_t, {rea}_{t}) \), which is generated by the following representation:
\begin{small}
\begin{equation}
\begin{aligned}
F_{{t}}^{act} = LLM(\varphi^{{act}}(s_t, F_{t}^{sum}, F_{t}^{ins}, {\lambda}_{base})),
\end{aligned}
\end{equation}
\end{small}where $\varphi^{act}(\cdot)$ is a prompt template for action making. This step is very important as it turns the gathered insights into a specific action. In the context of RTBAgent, this action is deciding on the bidding price for an ad impression. The final bidding price \( b^i_t \) for the current impression is calculated using the following formula:
\begin{small}
\begin{equation}
\begin{aligned}
 b^i_t = v^i_t \cdot {\lambda}_{base} \cdot (1 + a_t),
\end{aligned}
\end{equation}
\end{small}where \( b^i_t \) is the bidding price for the impression opportunity $i$ at time step $t$. This formula reflects the dynamic nature of the bidding process, allowing the agent to adjust its bids in real-time based on current analysis and historical data, thus optimizing its bidding strategy. 

\subsubsection{Daily Reflection}
Post the daily bidding cycle, RTBAgent engages in a process of introspection, encapsulated by \( F_{{t}}^{ref} \), to consolidate and reflect upon the day's decisions and their outcomes. This reflective process is integral to the continuous improvement of the agent's strategic acumen:
\begin{small}
\begin{equation}
\begin{aligned}
F_{{t}}^{ref} = \sum_{{i=bid,env,ref}} LLM(\varphi^{{ref}}(m^i_{t})),
\end{aligned}
\end{equation}
\end{small}where $\varphi^{ref}(\cdot)$ is a prompt template for daily reflection. This cyclical reflection ensures that the agent learns from its experiences, thereby refining its approach for enhanced performance in subsequent bidding endeavors.

The complete RTBAgent full-time bidding framework is provided in Algorithm  \ref{alg:rtbagent_framework}.

\begin{algorithm}[]
   \caption{Workflow of RTBAgent}
   \label{alg:rtbagent_framework}

   \begin{algorithmic}[1]
        \STATE Set Tools $\mathcal{H}$ and initialize Memory $\mathcal{M}$;
        \STATE Set prompt template $\varphi^{sum}(\cdot)$, $\varphi^{ins}(\cdot)$, $\varphi^{act}(\cdot)$, $\varphi^{ref}(\cdot)$;
        \STATE Set $K$ as the number of duration days;
        \STATE Set $T$ as the number of time steps for each day;

        \STATE Set budget list $B_{list}$ = $\{B_{1}, B_{2}, \dotsc,  B_{k}\}$
        \FOR{$k=1$ to $K$}
        \STATE Obtain allocated budget $B_k$;
        \FOR{$t=1$ to $T$}
        \STATE Observe state $s_t$;
        \STATE Obtain impression feature vector $X_t$;
        \STATE Obtain gathered information $F^{sum}_t$ via Eq.(7);
        \STATE Use $H$ to obtain $\lambda_{base}$ and  value estimation $V_t$ ;
        \STATE Obtain insight reasoning result $F^{ins}_t$ via Eq.(9);
        \STATE Obtain action making result  $F^{act}_t$ via Eq.(10);
        \STATE Get action \( a_t \) and reason \( {rea}_{t} \)  from  $F^{act}_t$;
        \STATE Bid each impression via Eq.(11);

        \STATE Store $s_t$, $F^{sum}_t$ and $F^{act}_t$  to $\mathcal{M}$;

        \IF{$t==T$}
        \STATE Obtain daily reflection $F^{ref}_t$ via Eq.(12);
        \STATE Store  $F^{ref}_t$ to $\mathcal{M}$;
        \ENDIF

        \ENDFOR

        \ENDFOR

   \end{algorithmic}

\end{algorithm}

\vspace{-0.4cm}

\section{Experiments}
\subsection{Datasets}  
We conduct a detailed study on the performance of RTBAgent using the iPinYou dataset~\cite{liao2014ipinyou}. The iPinYou dataset is provided by iPinYou Corporation, a prominent e-commerce advertising technology company in China. The dataset includes real-time bidding advertising data over a 10-day period in 2013, covering nine different advertising campaigns. Specifically, it contains 19.5 million ad displays, 14,790 clicks, and a total advertising cost of 16,000 RMB. These data not only portray the market environment, but also provide a complete path of user response from the advertisers' perspective. The records in the dataset are organized with each line representing three types of information: auction and ad features, auction winning price, and user click feedback on ad displays. Additionally, all monetary values are in RMB, corresponding to the cost-per-thousand-impressions (CPM) pricing model. The test data is derived from the final three days of each campaign, while the remaining data is used for training, as reported by the data publisher \cite{liao2014ipinyou}. 

\subsection{Evaluation Procedure}  
In our study, we specifically focus on the number of clicks as the KPI for evaluation. For each advertising campaign, we allocate the budget on a daily basis and divide each day into 24 time steps, representing each hour. The bidding model iterates the test dataset using the same CTR estimator. For each bidding request, the strategy generates a bid price that does not exceed the current budget. If this bid price is equal to or higher than the market price, the advertiser wins the auction, incurs the market price as a cost, and gains user clicks as a reward. Then, the remaining auction quantity and budget are updated. In order to prevent a situation where the budget is too high, and all bids are won, the budget limit cannot exceed the total historical cost of the test data. To examine the performance of bidding strategies under different budget constraints, we evaluate using three different budgets: 1/2, 1/8, and 1/32 of the total budget. 

\begin{table}[h]
  \centering
  \caption{Performance comparison of all methods in terms of actual click counts under different budgets.}
    \begin{tabular}{cccc}
    \toprule
    Model & 1/2   & 1/8   & 1/32 \\
    \midrule
    MCPC  & 1,779 & 989   & 298 \\
    LIN   & 2,200 & 1,202 & 744 \\
    LP    & 2,211 & 1,182 & 765 \\
    ORTB  & 2,222 & 1,192 & 746 \\
    RLB   & 2,223 & 1,227 & 741 \\
    DRLB  & 2,264 & 1,231 & 778 \\
    USCB  & 2,268 & 1,227 & 762 \\
    DiffBid  & 2,275 & 1,235 & 793 \\
    RTBAgent(ours) & \textbf{2,281} & \textbf{1,240} & \textbf{795} \\ \bottomrule
    \end{tabular}%
  \label{tab:compbaseline}%
  \vspace{-0.5cm}
\end{table}

\subsection{Implementation Details}
We utilize a single NVIDIA RTX A6000 GPU to train the RL model in our benchmark tests to ensure consistency in the training environment. All comparison schemes employed FM for CTR estimation. The FM model includes a linear layer for mapping features to the output space, a bias term for baseline prediction adjustment, and a feature embedding layer for capturing feature interactions. Finally, the sigmoid function converts the output into a probability form to predict the likelihood of user clicks. During the testing phase, the derivation of basic factor $\lambda_{base}$ by RTBAgent, based on expert strategies, is exclusively derived from the train set.     

\begin{table}[htbp]
\tabcolsep=0.09cm
  \centering
  \caption{The improvement in click counts brought by different expert strategies, among which the use of LLMs is Llama-3-8B-Instruct. The number in parentheses represents the performance improvement (\%) compared to the corresponding expert strategy. }
    \begin{tabular}{cccc}
    \toprule
    Budget & MCPC+ & LIN+  & LP+ \\
    \midrule
    1/2   & 1,796 (+0.96\%) & 2,216 (+0.73\%) & 2,265 (+2.44\%) \\
    1/8   & 1,070 (+8.19\%) & 1,223 (+1.75\%) & 1,252 (+5.92\%) \\
    1/32  & 336 (+12.75\%) & 783 (+5.24\%) & 796 (+4.05\%) \\ \bottomrule
    \end{tabular}%
  \label{tab:compstrategy}%
  \vspace{-0.5cm}
\end{table}%

\subsection{Overall Comparison} 
\label{sec:overallcomparison}
To verify the effectiveness of our model on real-world dataset, we compare it with the following competitive methods.

\textbf{MCPC} determines the maximum effective cost-per-click(CPC) for each advertising campaign by dividing the cost by the number of clicks in the training data, using this value as a parameter for bidding.

\textbf{LIN} is a linear bidding method where the bid value is linearly proportional to the estimated CTR \(\theta_e\). The bid for a single impression is formalized as \(\frac{b_0}{\theta_0}\theta_e\), where \(\theta_0\) is the average CTR in the train set, and \(b_0\) is an adjustment parameter. The optimal parameter combination \(\frac{b_0}{\theta_0}\) is selected for the bidding process.

\textbf{LP} models the problem as a linear programming issue. It directly solves for optimal offline parameters based on historical data to maximize the objective and uses these parameters for bidding.

\textbf{ORTB}~\cite{zhang2014optimal} regards RTB as a functional optimization problem and derives a bidding function.

\textbf{RLB}~\cite{cai2017real} uses a model-based RL method to maximize the total value of winning impressions within budget constraints.

\textbf{DRLB} utilizes the Deep Q-Network algorithm to train the optimal action strategy. 

\textbf{USCB} abstracts the core requirements of constrained bidding. It employs more efficient strategy search methods to achieve accelerated convergence.

\textbf{DiffBid}\cite{10.1145/3637528.3671526} leverages generative diffusion modeling to generate optimal bidding trajectories based on return correlations. According to \cite{10.1145/3637528.3671526}, we use USCB to generate training data for its non interactive learning through bidding activities.

\begin{table}[htbp]
\tabcolsep=0.10cm
  \centering
  \caption{The impact of using different LLMs in terms of actual click counts under different budgets, with the expert strategy used being LP. The numbers in parentheses indicate the increase in clicks compared to LP.}
    \begin{tabular}{cccc}
    \toprule
    LLMs   & 1/2   & 1/8   & 1/32 \\
    \midrule
    GPT-3.5-Turbo-1106 & 2,281 (+70) & 1,240 (+58) & 795 (+30) \\
    Glm-4-Air & 2,242 (+31) & 1,208 (+26) & 776 (+11) \\
    GPT-4o mini & 2,255 (+44) & 1,241 (+59) & 788 (+23) \\
    Llama-3-8B-Instruct & 2,265 (+54) & 1,252 (+70)  & 796 (+31) \\  
    Baichuan3-Turbo & 2,258 (+47) & 1,224 (+42) & 782 (+17) \\ 
    Yi-large & 2,220 (+9) & 1,245 (+63) & 792 (+27) \\
    \bottomrule
    \end{tabular}%
  \label{tab:compllms}%
    \vspace{-0.3cm}

\end{table}%

In our comparative analysis of performance on the iPinYou dataset under various budget scenarios, RTBAgent consistently outperformed conventional bidding methods and RL approaches, as shown in Table \ref{tab:compbaseline}. RTBAgent significantly improved the key metric of click number compared to traditional methods such as MCPC, LIN, LP, and ORTB. These traditional methods underperformed in budget management, primarily due to their lack of flexibility in adapting to market volatility, often resulting in early budget depletion. In contrast, RTBAgent’s advanced multi-memory retrieval system and two-step decision-making process allow for dynamic adjustments in bidding strategies, optimizing budget usage and enhancing advertising campaign effectiveness. When compared with RL models like RLB, DRLB and USCB, RTBAgent demonstrated superior adaptability and stability. Although model-based RL methods are generally well-suited to dynamic environments, they showed inconsistent performance under varying budget constraints, suggesting challenges in model training and strategic refinement.

Overall, the comprehensive evaluation metrics confirm RTBAgent’s outstanding performance, validating its effectiveness in advertising bidding and underscoring its robustness across different budgetary conditions.

\begin{figure*}[!t]
    \centering
    \includegraphics[width=0.99\linewidth]{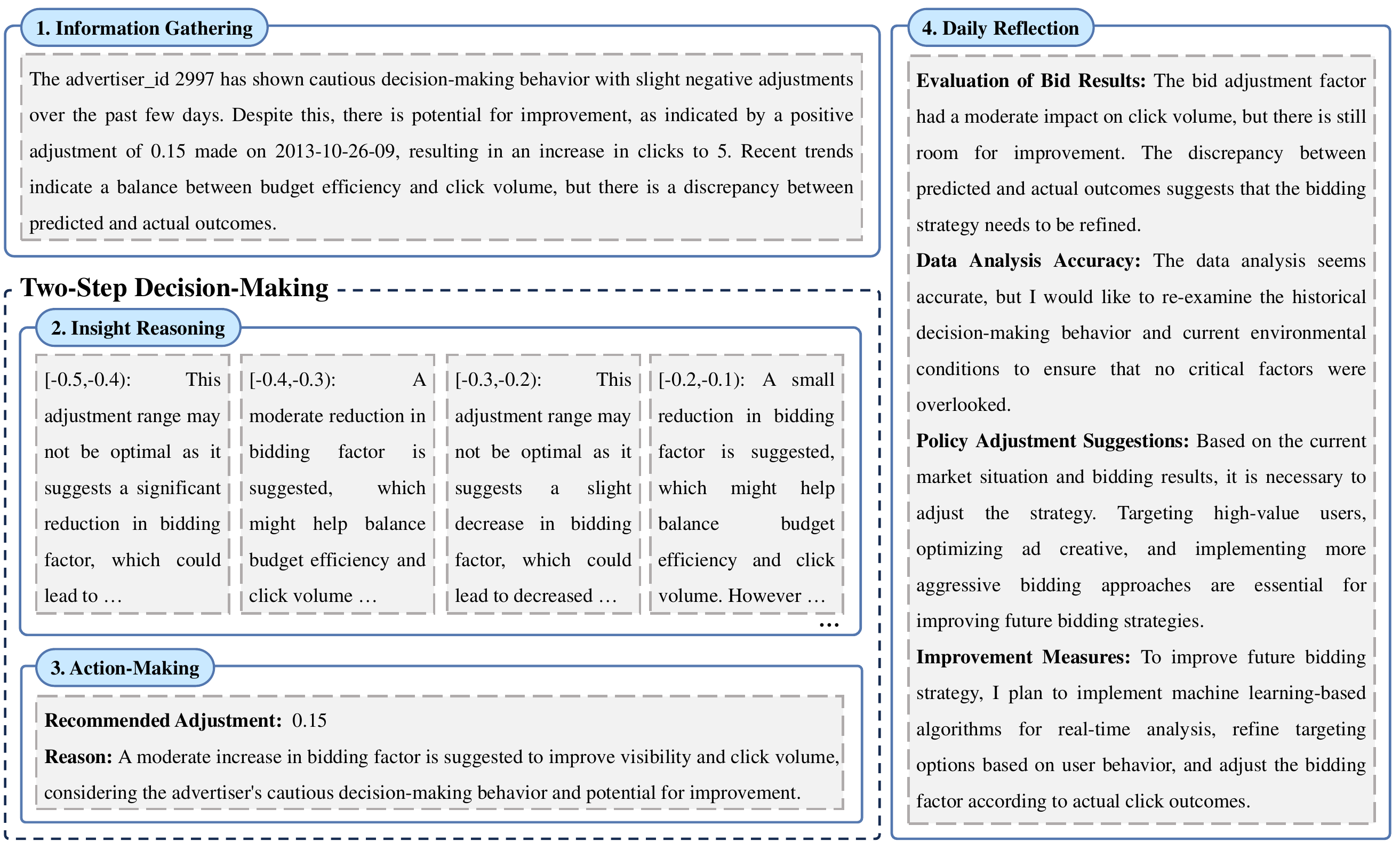}
    \caption{An example of RTBAgent using LLMs to assist reasoning process.}
    \label{fig:example}
    \vspace{-0.2cm}
\end{figure*}

\subsection{Ablation Study}
In Table \ref{tab:compstrategy}, we introduce three models, MCPC+, LIN+, and LP+, to demonstrate the performance enhancement achieved by RTBAgent with three expert strategies. Specifically, MCPC+, LIN+, and LP+ correspond to RTBAgent based on three different expert strategies, respectively: MCPC, LIN, and LP. Regardless of which method is used, the performance of RTBAgent on the test set consistently improved based on the original performance. It is observed that the stronger the expert strategy performance provided, the better the final results. 

\begin{figure}[htbp]
    \centering
    \includegraphics[width=0.99\linewidth]
    {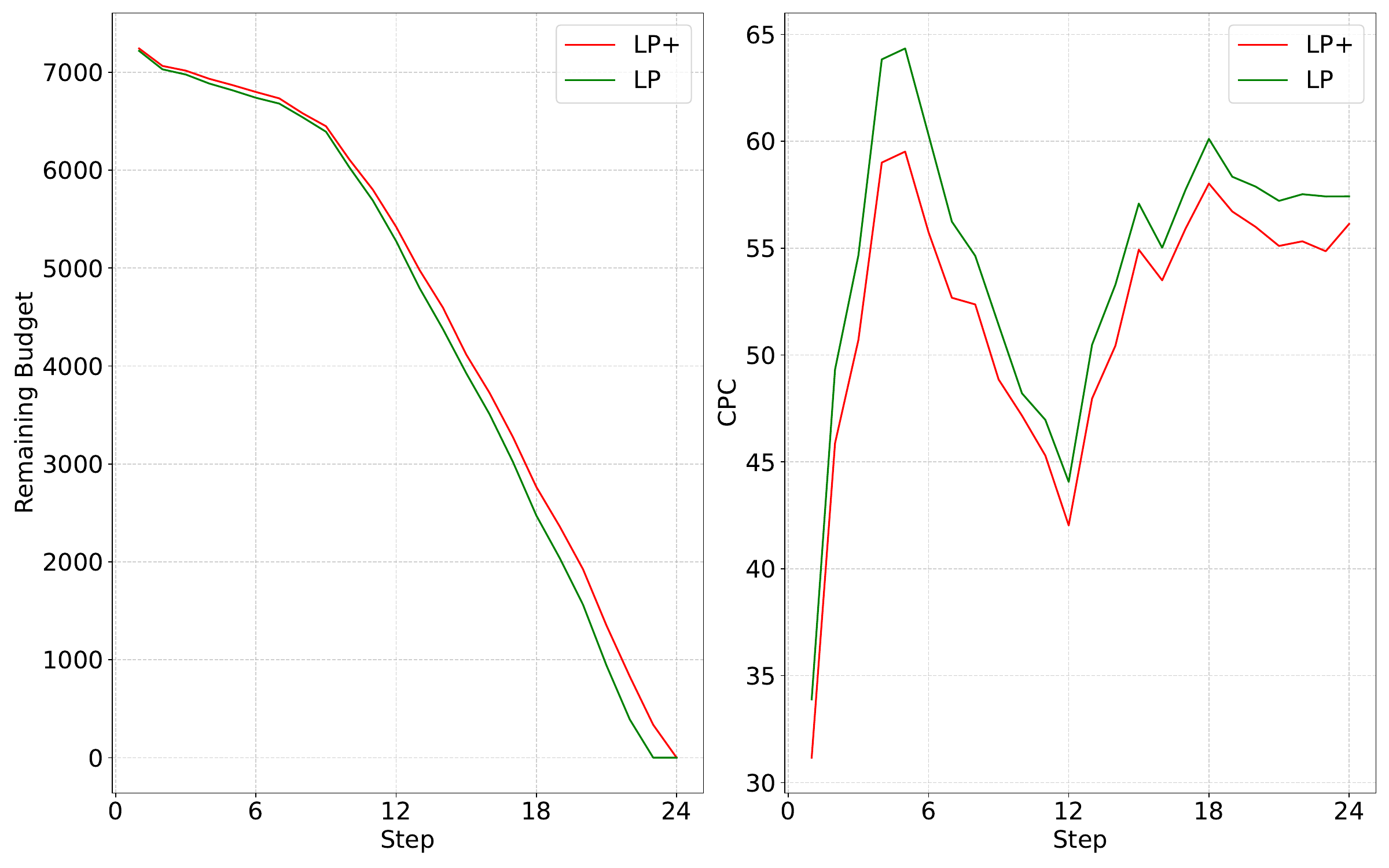}
    \caption{Comparison of the remaining budget and CPC between FP+ and FP at each step for advertiser 1458 on June 13th, 2013. The use of LLMs is Llama3-8B-Instruct.}
    \label{fig:1458-2013-06-14}
    \vspace{-0.5cm}
    
\end{figure}

As shown in Figure \ref{fig:1458-2013-06-14}, LP consumed too much budget in the early stages and did not anticipate the benefits of spending the budget in the later stages. On the contrary, LP+ can better control the expenditure of the budget, allowing for the purchase of high-quality clicks with a surplus budget in the second half of the process. In addition, LP+'s CPC has always been lower than the expert strategy throughout the process to ensure that more clicks are obtained within the specified budget. This suggests that RTBAgent effectively utilizes the guidance knowledge of expert strategies, leading to enhanced performance through insight into the environment and interaction.

\begin{table}[htbp]
  \centering
  \caption{Validation of the expert strategy and two-step decision-making effectiveness under different budgets, in which the expert strategy is LP, and the use of LLMs is Llama-3-8B-Instruct.}
    \begin{tabular}{ccc|ccc}
    \toprule
    Strategy & Insight & Action
 & 1/2   & 1/8   & 1/32 \\
    \midrule
         &      & {\Checkmark}      & 1,796  & 634   & 227 \\
    {\Checkmark}      &      & {\Checkmark}      & 2,194  & 1,186  & 779 \\
    {\Checkmark}      & {\Checkmark}      & {\Checkmark}      & \textbf{2,265} & \textbf{1,252} & \textbf{796} \\ \bottomrule
    \end{tabular}%
  \label{tab:com2step}%
    \vspace{-0.2cm}

\end{table}%

The performance of RTBAgent, utilizing distinct scale base models for real-time bidding tasks, is examined in Table \ref{tab:compllms}. Comparative analysis reveals that RTBAgent consistently surpasses alternative strategies across all budget allocation ratios. This indicates RTBAgent's superior performance regardless of model scale, from larger, more capable models to smaller, more efficient ones. The agent demonstrates adaptability and scalability, capable of dynamically adapting to the model size in response to business requirements and resource limitations, thus achieving the best bidding outcomes.

In Table \ref{tab:com2step}, we analyze the effectiveness of expert strategies and two-step decision-making (including insight reasoning and action making). Specifically, the action directly utilizes output optimal scaling factor by LLMs for situations without expert strategy. By comparing rows one and two of Table \ref{tab:com2step}, we can find that incorporating expert knowledge boosts model performance by utilizing historical market information. Furthermore, row three highlights the superiority of a two-step decision-making model over direct action approaches regarding click count increases, emphasizing the value of detailed market analysis and flexible bidding strategies.

\subsection{Visualization and Analysis of Performance}
Figure \ref{fig:example} demonstrates how RTBAgent utilizes LLMs to enhance its reasoning process during real-time bidding, specifically for advertiser 2997 on the last step of October 27th, 2013. The figure illustrates the core workflow, starting with information gathering, where RTBAgent analyzes historical bidding data, identifying that a reasonable increase in the bidding factor has effectively improved visibility and click volume. This data analysis is the foundation for the next stage, a two-step decision-making process. In the first step, RTBAgent performs insight reasoning by analyzing current market conditions and historical data to extract strategic insights, such as recommending a 0.15 adjustment to the bid factor based on past performance. In the second step, RTBAgent translates these insights into actionable decisions by adjusting the bid price, aiming to optimize performance while maintaining stability. After each bidding cycle, RTBAgent engages in daily reflection, where it evaluates the outcomes of its decisions, compares expected results with actual performance, and updates its strategy accordingly. This iterative process allows RTBAgent to continuously improve its bidding approach, adapting to market dynamics and making more informed, strategic decisions in future cycles. The visualization in Figure \ref{fig:example}, along with the data presented in Table \ref{tab:com2step}, underscores RTBAgent's capability to refine the decision-making process in RTB scenarios. The detailed analysis of a single hour for advertiser 2997 on October 27th, 2013, demonstrates RTBAgent’s ability to learn from daily operations and continuously improve its bidding strategies.

Furthermore, we invited 10 experts in the advertising field to rate the generated decisions and reasons on three levels: -1, 0, and 1. Finally, RTBAgent received a positive rating of 97\%, indicating that the proposed RTBAgent's output is convincing enough.
\subsection{Discussion of Reasoning Costs and Benefits}

In advertising bidding systems, methods that rely on RL or rules are typically triggered at longer intervals, such as every 15 minutes or more. Despite the inference time for LLMs being slightly longer than these traditional methods, it is still adequate to satisfy the typical latency requirements of advertising services in practice. As shown in Table \ref{tab:compllms}, our framework outperforms other baselines when using LLMs of various scales, ranging from big to small, whether they are open-source LLMs or closed-source LLMs. This performance advantage allows advertisers to select the LLMs that best fit their deployment needs, reducing the concern over inference cost. Additionally, as demonstrated in Figure \ref{fig:example}, our framework delivers not only reliable results but also provides data-driven insights and decision-making rationale, which is invaluable for operations teams when assessing the effectiveness of advertising campaigns and refining future strategies. Consequently, our proposed solution not only fulfills the fundamental requirements for inference time and cost but also generates direct economic benefits and indirect operational advantages.

\section{Conclusions}

In this paper, we introduce RTBAgent, an effective agent that first utilizes LLMs to enhance advertising auctions in RTB. Specifically, we innovatively propose a two-step decision-making process that integrates CTR estimation model, expert strategy knowledge, multi-memory retrieval system, and self-reflection, providing accurate simulation and real-time decision support for bidding scenarios. The extensive experimental results confirm that RTBAgent exhibits better adaptability and interpretability than traditional rule-based and RL methods in highly dynamic and unstable bidding environments. Our work contributes to developing a novel paradigm that attempts to explore the application of LLM-based integrated intelligence in RTB and spark related discussions. In future research, in order to better fit the competitiveness of the online advertising market, we will focus on studying the application of multi-agent systems based on LLMs in RTB. It can be foreseen that multi-agent systems will become more complex and effective.

\section{Limitations}
Our approach does have some limitations. Firstly, while LLM-based methods have achieved improvements in effectiveness, the response time of LLM-based bidding systems is not as swift as desired. Employing models with smaller parameter sizes, such as 1B or 3B, appears to be a promising direction. Secondly, current LLMs have not yet encompassed a richer set of bidding knowledge. Utilizing strategies like RAG might lead to greater enhancements, which is on our agenda for the future. Thirdly, we have deployed this system into a real-world advertising bidding environment, and we plan to disclose the revenue in the future.

\section{Acknowledgments}
This work is supported in part by the National Natural Science Foundation of China (62372187), in part by the National Key Research and Development Program of China (2022YFC3601005) and in part by the Guangdong Provincial Key Laboratory of Human Digital Twin (2022B1212010004).

\onecolumn
\begin{multicols}{2}
    \bibliographystyle{ACM-Reference-Format}
    \bibliography{sample-authordraft}
\end{multicols}

\section{appendix}
\lstset{
 columns=fixed,       
 numbers=left,                                        
 numberstyle=\tiny\color{gray},                       
 frame=lines,                                        
 backgroundcolor=\color[RGB]{255,255,255},            
 keywordstyle=\color[RGB]{40,40,255},                 
 numberstyle=\footnotesize\color{darkgray},           
 commentstyle=\color[RGB]{0,0,0},                
 stringstyle=\rmfamily\slshape\color[RGB]{128,0,0},   
 showstringspaces=false,                              
 language=python,                                        
 title=\lstname,
 framexleftmargin = 6mm,
 basicstyle=\small\ttfamily,
}

\subsection{Partial Display of Prompt Template for RTBAgent}
Due to space limitations, only a portion of the prompt templates for RTBAgent is displayed. We will organize and publish the remaining codes and prompt templates soon.
\begin{center}

\noindent\textit{Profile Definition Template of Insight Reasoning.}
\vspace{+1em}
\end{center}
\vspace{-1em}

\begin{lstlisting}[language=Python]
"""
# YOUR ROLE
- You are a senior data analyst specializing in in-depth research and strategy development in the field of real-time bidding (RTB) advertising placement.
- You use advanced data analysis tools and algorithms to guide advertisers to gain an advantage in fierce market competition.
- Your goal is to maximize the number of clicks through data analysis and algorithm adjustments given the budget for the day.
- You need to regularly adjust the bidding factor based on historical decision-making, current environmental conditions, and algorithmic bidding recommendations.

# CONTEXT
{history}

# THE REFERENCE GIVEN BY THE BIDDING ALGORITHM FOR ADVERTISER
{bidding_reference}
# ENVIRONMENT STATUS
{environment_status}
# NOW YOUR ACTION IS
The bidding factor for this period = the bidding factor given by the algorithm * (1+adjustment)
Now, you need to analyze the advantages and disadvantages of each "adjustment range" based on historical decisions, current environmental conditions, and algorithm suggestions. 
The selection space for "adjustment range" is from {{[-0.5,-0.4), [-0.4,-0.3), [-0.3,-0.2), [-0.2,-0.1), [-0.1,0.0), [0.0,0.1), [0.1,0.2), [0.2,0.3), [0.3,0.4), [0.4,0.5]}}.
While making your analysis, consider the following:
1. **Historical Performance**: Adjustments that have previously optimized budget usage and improved click volume without significantly impacting the win rate are valuable.
2. **Exploration of New Adjustments**: Exploring new adjustment ranges, especially positive adjustments range like [0.0,0.1), [0.1,0.2), and [0.2,0.3), can potentially uncover more effective strategies. Increasing the bid may improve visibility and click volume, particularly in competitive environments.
3. **Balancing Stability and Innovation**: Strive to balance between maintaining strategies that have shown consistent performance and exploring new adjustments. A mix of historical strategy and new exploration, with a focus on positive adjustments, could provide a balanced approach to ensure cost efficiency, maximize clicks, and adapt to market changes.
Please output a JSON in the following format for all analyses:
```json
{{
    "adjustment range for [-0.5,-0.4)": str = xx,
    "adjustment range for [-0.4,-0.3)": str = xx,
    "adjustment range for [-0.3,-0.2)": str = xx,
    "adjustment range for [-0.2,-0.1)": str = xx,
    "adjustment range for [-0.1,0.0)": str = xx,
    "adjustment range for [0.0,0.1)": str = xx,
    "adjustment range for [0.1,0.2)": str = xx,
    "adjustment range for [0.2,0.3)": str = xx,
    "adjustment range for [0.3,0.4)": str = xx,
    "adjustment range for [0.4,0.5]": str = xx
}}
"""
\end{lstlisting}

\begin{center}
\vspace{+1em}

\noindent\textit{Profile Definition Template of History Bidding Summary.}
\end{center}
\vspace{-0.5em}

\begin{lstlisting}[language=Python]
"""
You are an advanced history summary tool specializing in big data insights. Your task is to analyze recent trends and evaluate the effectiveness of decision-making behaviors over the selected time window.
Please summarize the following information with a focus on recent changes, highlighting any new patterns, shifts, or significant deviations from past behavior. Pay particular attention to the current environment information provided at the end of this document. Use concise language to ensure the summary is clear and actionable.
Summarize and return the output strictly in the following JSON format, without any additional text or explanations:
```json
{{
    "summary": str = xx
}}
```
"""
\end{lstlisting}

\end{document}